\def\year{2020}\relax
\documentclass[letterpaper]{article} % DO NOT CHANGE THIS
\usepackage{aaai20}  % DO NOT CHANGE THIS
\usepackage{times}  % DO NOT CHANGE THIS
\usepackage{helvet} % DO NOT CHANGE THIS
\usepackage{courier}  % DO NOT CHANGE THIS
\usepackage[hyphens]{url}  % DO NOT CHANGE THIS
\usepackage{graphicx} % DO NOT CHANGE THIS
\usepackage{amsmath}
\usepackage{amssymb}
\usepackage{booktabs}
\usepackage{mathrsfs}
\usepackage{boldline}
\usepackage{multirow}
\usepackage{subcaption}

\title{Unsupervised Neural Dialect Translation with \\ Commonality and Diversity Modeling}
\author{Yu Wan\thanks{Equal contribution.}~~~~~Baosong Yang$^*$~~~~~Derek F. Wong\thanks{Corresponding author.}~~~~~Lidia S. Chao~~~~~Haihua Du~~~~~Ben C.H. Ao\\
  NLP$^2$CT Lab, Department of Computer and Information Science, 
  University of Macau\\
  {\tt nlp2ct.\{ywan,baosong,duhaihua,benao\}@gmail.com, \{derekfw,lidiasc\}@um.edu.mo}}

\begin{document}

\maketitle
 \begin{abstract}
As a special machine translation task, dialect translation has two main characteristics: 1) lack of parallel training corpus; and 2) possessing similar grammar between two sides of the translation. 
In this paper, we investigate how to exploit the commonality and diversity between dialects thus to build unsupervised translation models merely accessing to monolingual data.
Specifically, we leverage pivot-private embedding, layer coordination, as well as parameter sharing to sufficiently model commonality and diversity among source and target, ranging from lexical, through syntactic, to semantic levels.  
In order to examine the effectiveness of the proposed models, we collect 20 million monolingual corpus for each of Mandarin and Cantonese, which are official language and the most widely used dialect in China. 
Experimental results reveal that our methods outperform rule-based simplified and traditional Chinese conversion and conventional unsupervised translation models over 12 BLEU scores. 

\end{abstract}

\section{Introduction}
\label{sec.Intro}

\begin{figure}
    \centering
    \includegraphics[keepaspectratio, width=0.45\textwidth]{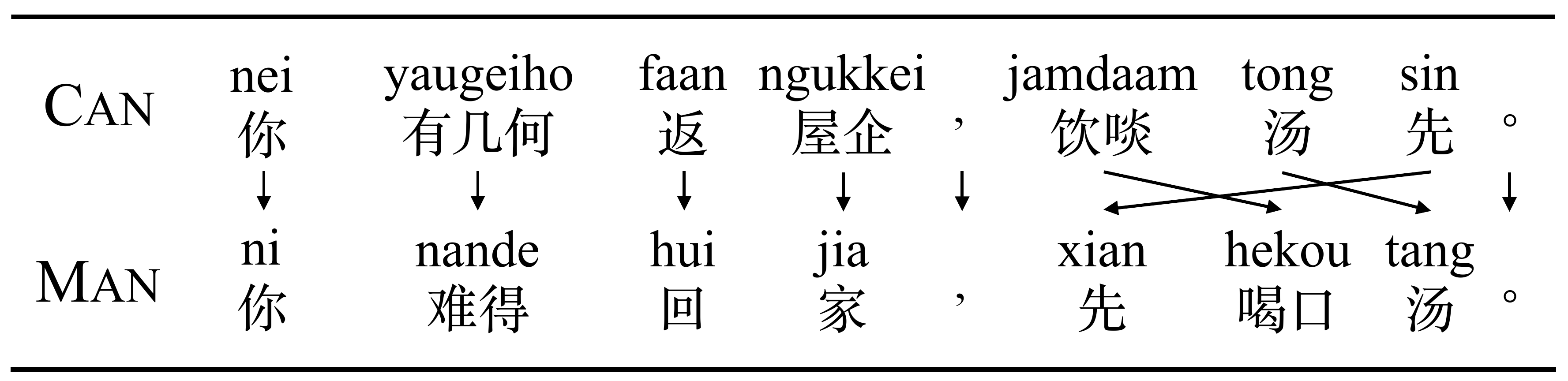}
    \caption{An example of \textsc{Can}-\textsc{Man} translation.}
    \label{Fig.Examples}
\end{figure}

Dialect refers to a variant of a given language, which could be defined by factors of regional speech patterns, social class or ethnicity~\cite{lyons1981language}. Except for pronunciation, a dialect is also distinguished by its textual expression~\cite{wong2018register}. For instance, Mandarin (\textsc{Man}) and Cantonese (\textsc{Can}) are the official language and the most widely used dialect of China, respectively~\cite{lee1998cancorp}. As seen in Fig.~\ref{Fig.Examples}, although the sentences have absolutely same semantic meaning, they have distinct attributes with respect to the expression on text. Correspondingly, in this task we attempt to build automatic translation system for dialects.

An intuitive way is to leverage advanced machine translation systems which have recently yielded human-level performance with the use of neural networks~\cite{Chen:2018:ACL,li2018multi}.  Nevertheless, contrast with traditional machine translation, there are two main challenges in dialect translation. First, the success of supervised neural machine translation depends on large-scale training parallel data, while dialect translation is not equipped such kind of prerequisite. This makes our task fall into unsupervised learning category~\cite{artetxe2018unsupervised,lample2018unsupervised,lample2018phrase}. Second, dialects are closely related and, despite their differences, often share similar grammar, e.g. morphology and syntax~\cite{chambers_trudgill_1998}. The extraction of {\em commonality} is beneficial to unsupervised mapping~\cite{lample2018word} and model robustness~\cite{firat2016multi}, in the meanwhile, preserving the explicit {\em diversity} plays a crucial role in our dialect translation. Consequently, it is challenging to balance the commonality and diversity for dialect translation thus to improve its performance.

We approach the mentioned problems by proposing unsupervised neural dialect translation model, which is merely trained using monolingual corpus and sufficiently leverage commonality and diversity of dialects.
Specifically, we train an advanced NMT model~\textsc{Transformer}~\cite{vaswani2017attention} with denoising reconstruction~\cite{vincent2008extracting} and back-translation~\cite{sennrich2016improving}, which aim at building common language model and mapping different attributes, respectively. 
We introduce several strategies into translation model for balancing the commonality and diversity: 1) parameter-sharing that forces dialects to share the same latent space; 2) pivot-private embedding which models similarities and differences at lexical level; and 3) layer coordination which enhances the interaction of features between two sides of translation.

In order to evaluate the effectiveness of the proposed model, we propose monolingual dialect corpus which consists of 20 million colloquial sentences for each of~\textsc{Man}\footnote{For simplification, we regard official language as a dialect.} and~\textsc{Can}. 
The sentences are extracted from conversations and comments in forums, social medias as well as subtitles, and carefully filtered during data preprocessing.\footnote{Our codes and data are released at:~\url{https://github.com/NLP2CT/Unsupervised_Dialect_Translation}.} 
Empirical results on two directions of \textsc{Man}-\textsc{Can} translation task demonstrate that the proposed model significantly outperforms existing unsupervised NMT~\cite{lample2018phrase} with even fewer parameters. The quantitative and qualitative analyses verified the necessity of commonality and diversity modeling for dialect translation. 

\section{Preliminary}
\label{sec.Preliminary} 
Neural machine translation (NMT) aims to use a neural network to build a translation model, which is trained to maximize the conditional distribution of sentence pairs~\cite{bahdanao2014neural,sennrich2016neural,vaswani2017attention}.
Given a source sentence $\mathbf{X}=\{\mathbf{x}_1,\cdots,\mathbf{x}_{I}\}$, conditional probability of its corresponding translation $\mathbf{Y}=\{\mathbf{y}_1,\cdots,\mathbf{y}_{J}\}$ is defined as: 
\begin{align}
    \mathbf{P}(\mathbf{Y} | \mathbf{X})&=\prod \limits_{j=1}^{|{J}|}\mathbf{P}(\mathbf{y}_j|\mathbf{Y}_{<j},\mathbf{X};\theta),
\end{align}
where $\mathbf{y}_j$ indicates the $j$-th target token. $\theta$ denotes the parameters of NMT model, which are optimized to minimize the following loss function over the training corpus $\mathbf{D}$:
\begin{align}
\label{eq:lossbase}
    \mathcal{L} &= \mathbb{E}_{(\mathbf{X},\mathbf{Y})\sim \mathbf{D}}[-\log \mathbf{P} (\mathbf{Y} | \mathbf{X};\theta)]
\end{align}

Such kind of auto-regressive translation process is generally achieved upon the encoder-decoder framework~\cite{sutskever2014sequence}. Specifically, the inputs of encoder $\textbf{S}^0$ and decoder $\textbf{T}^0$ are obtained by looking up source and target embeddings according to the input sentences $\textbf{X}$ and $\textbf{Y}$, respectively:
\begin{align}
    \textbf{S}^0 &= {\rm{Emb}}_{src}(\textbf{X}) &  \in \mathbb{R}^{I \times d} \label{eq:embs}\\
    \textbf{T}^0 &= {\rm{Emb}}_{trg}(\textbf{Y}) &  \in \mathbb{R}^{J \times d} \label{eq:embr}
\end{align} 
where $d$ indicates the dimensionality. The encoder is composed of a stack of $N$ identical layers. Given the input layer $\mathbf{S}^{n-1}\in \mathbb{R}^{I \times d}$, the output of the $n$-th layer can be formally expressed as:
\begin{align}
     {\rm\mathbf{S}}^{n} &={{\rm{Layer}}^{n}_{enc} ({\rm\mathbf{S}}^{n-1})} &  \in \mathbb{R}^{I \times d}
\end{align}
The decoder is also composed of a stack of $N$ identical layers. Contrary to the encoder which takes all the tokens into account, the decoder merely summarizes the forward representations in the input layer $\mathbf{T}^{n-1}\in {\mathbb{R}^{J \times d}}$ at each decoding step, since the subsequent representations are invisible. Besides, the generation process considers the contextual information of source sentence, by feeding the top layer of the encoder $\mathbf{S}^{N}$. Accordingly, the $j$-th representation in $n$-th decoding layer $\mathbf{T}^{n} = \{\mathbf{t}^{n}_1,\cdots,\mathbf{t}^{n}_J\}$ is calculated as:
\begin{align}
    \label{eq:dec}
    {\rm \mathbf{t}}^{n}_j &={\rm{Layer}}^{n}_{dec} (\mathbf{T}^{n-1}_{\leqslant j}, {\rm{Att}}^{{n}}
    (\mathbf{t}^{n-1}_j, \mathbf{S}^{N})) &  \in \mathbb{R}^{d}
\end{align}
where $\rm{Att}(\cdot)$ indicates the attention model~\cite{bahdanao2014neural} which has recently been a basic module to allow a deep learning model to dynamically select related representations as needed. Finally, the conditional probability of the $j$-th target word $\mathbf{y}_j$ is calculated using a non-linear function $\rm{Softmax(\cdot)}$:
\begin{align}
\mathbf{P}(\mathbf{y}_j|\mathbf{Y}_{<j},\mathbf{X};\theta) &= {\rm{Softmax}} ({\rm{Proj}} (\mathbf{t}^N_j))
\end{align}

\begin{figure*}[t!]
    \centering
    \hspace{-5pt}
    \begin{subfigure}[c]{\columnwidth}
    {
        \centering\includegraphics[keepaspectratio,height=0.65\columnwidth]{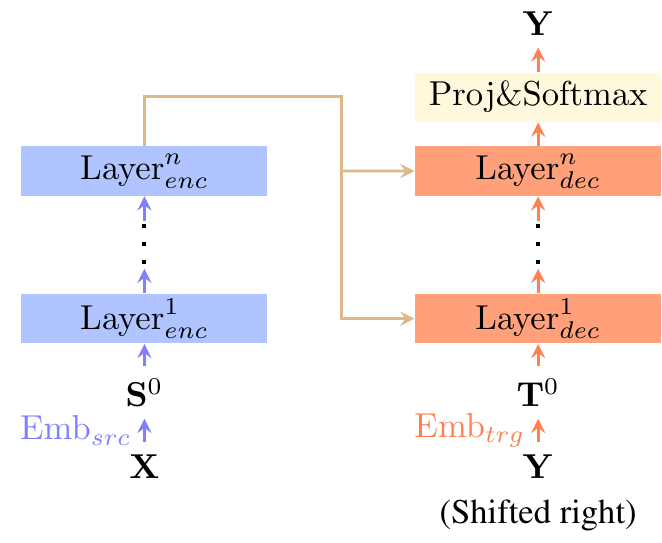}
        \caption{Conventional NMT model.}
    }
    \end{subfigure}
    \quad
    \hspace{15pt}
    \begin{subfigure}[c]{\columnwidth}
    {
        \centering\includegraphics[keepaspectratio,height=0.65\columnwidth]{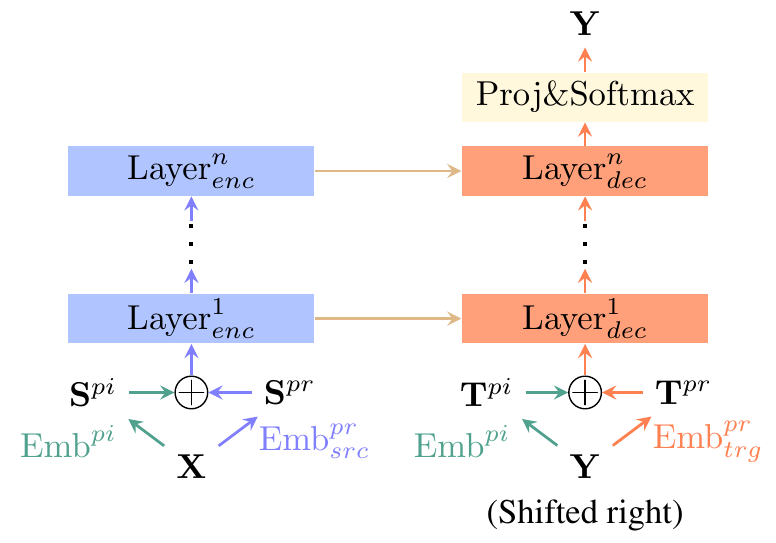}
        \caption{The proposed model.}
    }
    \end{subfigure}
    
    \caption{Illustration of (a) conventional NMT model and (b) the proposed model. As seen, we propose pivot-private embedding, which learns commonality (${\rm{Emb}}^{pi}$) and diversity (${\rm{Emb}}^{pr}_{src}$ and ${\rm{Emb}}^{pr}_{trg}$) at lexical level. Besides, the decoder attends to source representations layer by layer, rather than merely from the topmost layer.}
    \label{Fig.Model}
\end{figure*}

In this section, we propose unsupervised neural dialect translation. We first serve the dialect translation as an unsupervised learning task to tackle with the low-resource problem. Moreover, concerning the commonality and diversity between dialects, we introduce pivot-private embedding and layer coordinating to improve the dialect translation model.

\subsection{Dialect Translation with Unsupervised Learning}
\label{subsec.UnsupervisedNeuralMachineTranslation}

Despite the success of NMT over past years, the performance of a NMT model relies on large-scale parallel training corpus~\cite{sennrich2016improving,artetxe2018unsupervised}. As a low-resource translation task, dialect translation fails at leveraging conventional training strategy, since parallel resources are normally inaccessible. The scarcity of bilingual corpus leads to extraordinary challenging on building translation models for dialects. On the contrary, monolingual corpora is relatively easier to be collected. 
Partially inspired by recent studies on unsupervised NMT~\cite{lample2018unsupervised,artetxe2018unsupervised,lample2018phrase}, we propose to build dialect translation model with unsupervised learning which merely depends on monolingual data. 
Generally, most of the features with respect to dialects are similar, while only a few of the surface information is different. To this end, we propose to divide the training process into two parts: 1) commonality modeling which learns to capture general features of all dialects; and 2) diversity modeling which builds connections between different expressions. 

\paragraph{Commonality Modeling}
This procedure aims at offering our model the ability to extract the universal features of two dialects. Intuitively, the commonality modeling can be trained by reconstructing two dialects using one model.  \citeauthor{artetxe2018unsupervised}~\shortcite{artetxe2018unsupervised} and \citeauthor{lample2018unsupervised}~\shortcite{lample2018unsupervised} suggest that denoising autoencoding is beneficial to the language modeling. More importantly, it can avoid our model from severely copying the input sentence to the output. Contrary to~\citeauthor{artetxe2018unsupervised}~\shortcite{artetxe2018unsupervised} and \citeauthor{lample2018unsupervised}~\shortcite{lample2018unsupervised} who employ distinct model for each language, we train one model for both the two dialects, thus to encourage different dialects to be modeled under a common latent space. Consequently, the loss function is defined as:
\begin{align}
    \mathcal{L}_{com} =& \mathbb{E}_{\mathbf{X}\sim \mathbf{D}_X}[-\log \mathbf{P} (\mathbf{X} | \mathbf{X}^{noise};\theta)] +  \nonumber\\
    & \mathbb{E}_{\mathbf{Y}\sim \mathbf{D}_Y}[-\log \mathbf{P} (\mathbf{Y} | \mathbf{Y}^{noise};\theta)]
\end{align}
where $\mathbf{D}_X$ and $\mathbf{D}_Y$ are monolingual corpora for two dialects, $\mathbf{X}^{noise}$ and $\mathbf{Y}^{noise}$ denote noised inputs.\footnote{We add noises to inputs by swapping, dropping and blanking words following~\citeauthor{lample2018unsupervised}~\shortcite{lample2018unsupervised}, except that we swap two words rather than three, which shows better empirical results in our experiments.}
As seen, the two reconstruction models are shared with the same parameter $\theta$.
\paragraph{Diversity Modeling} Although there is marginal difference between dialects, the transfer of diversity is the key problem of dialect translation. Contrast to supervised NMT model which learns relevance between source and target using parallel data, dialect translation model fails to directly establish the functional mapping from source latent space to target one. An alternative way is to exploiting back-translation~\cite{sennrich2016improving,edunov-etal-2018-understanding}. Specifically, $\mathbf{X}$ and $\mathbf{Y}$ are first translated to their candidate translation $\mathbf{Y}^{bak}$ and $\mathbf{X}^{bak}$, respectively. 
The mapping of cross-dialect latent spaces can be learned by minimizing:
\begin{align}
    \mathcal{L}_{div} =& \mathbb{E}_{\mathbf{X}\sim \mathbf{D}_X}[-\log \mathbf{P} (\mathbf{X} | \mathbf{Y}^{bak};\theta)] +  \nonumber\\
    & \mathbb{E}_{\mathbf{Y}\sim \mathbf{D}_Y}[-\log \mathbf{P} (\mathbf{Y} | \mathbf{X}^{bak};\theta)]
\end{align}
Finally, the loss function in Equation~\ref{eq:lossbase} is modified as:
\begin{align}
    \mathcal{L} &= \lambda_{com}\mathcal{L}_{com} + \lambda_{div}\mathcal{L}_{div}
\end{align}
where $\lambda_{com}, \lambda_{div}$ are hyper-parameters balancing the importance of commonality and diversity modeling, respectively.

\subsection{Pivot-Private Embedding}
\label{subsec.PivotEmbedding}
An open problem in unsupervised NMT is the initialization of the translation model, which plays a crucial role in the iteratively training~\cite{lample2018unsupervised,artetxe2018unsupervised} and affects the final performance of the unsupervised learning~\cite{lample2018phrase}. 
For two languages with different vocabularies, an usual solution in recent studies is to map the same tokens which are then cast as seeds for aligning other words~\cite{artetxe2018unsupervised,lample2018unsupervised}. For example,~\citeauthor{artetxe2018unsupervised}~\shortcite{artetxe2018unsupervised} employ unsupervised bilingual word embeddings~\cite{artetxe2017learning}, while~\citeauthor{lample2018phrase}~\shortcite{lample2018phrase} utilize the representations of shared tokens~\cite{Mikolov2013Distributed} in different languages to initialize the lookup tables. 
Fortunately, dialect translation dispels this problem since most of tokens are shared among dialects. 
Therefore, we propose pivot and private embedding, in which, the former learns to share a part of the features while the latter captures the word-level characteristics in different dialects. 
\paragraph{Pivot Embedding}
Since vocabularies in different dialects are almost the same, we join monolingual corpora of two dialects and extract all the tokens in it. In order to build the connections between source and target, we assign pivot embedding with $d_s$ dimensions as the initial alignments:
\begin{align}
    \textbf{S}^{pi} &= {\rm{Emb}}^{pi}(\textbf{X}) &  \in \mathbb{R}^{I \times d_s} \\
    \textbf{T}^{pi} &= {\rm{Emb}}^{pi}(\textbf{Y}) &  \in \mathbb{R}^{J \times d_s}
\end{align} 
where the function of looking up embedding ${\rm{Emb}}^{pi}(\cdot)$ shares parameters across dialects. 
\paragraph{Private Embedding} Except the common features, there also exists differences between dialects. We argue that such kind of difference mainly lies in the word-level surface information. To this end, we introduce private embedding for each translation side to distinguish and maintain the characteristics in dialects:  
\begin{align}
    \textbf{S}^{pr} &= {\rm{Emb}}_{src}^{pr}(\textbf{X}) &  \in \mathbb{R}^{I \times (d-d_s)} \\
    \textbf{T}^{pr} &= {\rm{Emb}}_{trg}^{pr}(\textbf{Y}) &  \in \mathbb{R}^{J \times (d-d_s)}
\end{align} 
Contrary to pivot embedding, ${\rm{Emb}}_{src}^{pr}(\cdot)$ and ${\rm{Emb}}_{trg}^{pr}(\cdot)$ are assigned distinct parameters. Thus, the final input embedding in Equation~\ref{eq:embs} and \ref{eq:embr} are modified as:
\begin{align}
    \textbf{S}^0 &= \textbf{S}^{pr} \oplus \textbf{S}^{pi}  &  \in \mathbb{R}^{I \times d} \\
    \textbf{T}^0 &= \textbf{T}^{pr} \oplus \textbf{T}^{pi}  &  \in \mathbb{R}^{J \times d}
\end{align} 
where $\oplus$ is the concatenation operator. Note that, since each token has $d_s$ and $d-d_s$ dimensions for the associate pivot embedding and private embedding, the final input is still composed of $d$-dimensional vector. ${\rm{Emb}}^{pi}(\cdot)$,  ${\rm{Emb}}_{src}^{pr}(\cdot)$ and ${\rm{Emb}}_{trg}^{pr}(\cdot)$ are all pretrained, and co-optimized under the translation objective. In this way, we expect that pivot embedding can enhance the commonality of translation model, while private embedding raises the ability to capture the diversity of different dialects~\cite{liu19shared}. 

\subsection{Layer Coordination}
\label{subsec.CrossAttentionCoordinate}
Recent studies have pointed out that multiple neural network layers are able to capture different types of syntactic and semantic information~\cite{Peters:2018:NAACL,Li:2019:NAACL}. For example,~\citeauthor{Peters:2018:NAACL}~\shortcite{Peters:2018:NAACL} demonstrate that higher-level layer states capture the context-dependent aspects of word meaning while lower-level states model the aspects of syntax, and simultaneously exposing all of these signals is highly beneficial. 
To sufficiently interact these features, an alternative way is to perform attention from a decoder layer to its corresponding encoder layer, rather than merely from the topmost layer. Accordingly, the $n$-th decoding layer (Equation~\ref{eq:dec}) is changed to: 
\begin{align}
    \mathbf{t}^{n}_j &={{\rm{Layer}}^n_{dec}} (\mathbf{T}^{n-1}_{\leqslant j}, {{\rm{Att}}^n} (\mathbf{t}^{n-1}_j, \mathbf{S}^{n})) &  \in \mathbb{R}^{d}
\end{align}

This technique has been proven effective \cite{he2018layer,yang2019context,Hao:2019:NAACL} upon NMT tasks via shortening the path of gradient propagation, thus stabilizes the training of a extremely deep model. 
However, the improvements on traditional translation tasks become marginal when we apply layer coordination to the models with less than 6 layers \cite{he2018layer}. We attribute this to the fact that directly interacting lexical and syntactic level information between different languages affects the diversity modeling of them, since it forces the two languages to share the same latent space layer by layer. Different from prior studies, our work focuses on a pair of languages which have extremely similar grammar. We examine whether layer coordination is conductive to commonality modeling of dialects and the translation quality.

\section{Datasets}

In this section, we first introduce the \textsc{Can} and \textsc{Man} datasets collected for our experiments, then show adequate rudimentary statistical results upon training corpora.

\paragraph{Monolingual Corpora}
The lack of \textsc{Can} monolingual corpora with strong colloquial features is serious obstacle in our research.
Existing \textsc{Can} corpora, such as HKCanCor \cite{luke2015hong} and CANCORP \cite{lee1998cancorp}, all have the following shortcomings: 
1) they were collected in rather early years, the linguistic features of which vary from the current ones due to language evolution; and 
2) they are scarce for data-intensive unsupervised training.
Due to the fact that colloquial corpora possess more distinguished linguistic features of \textsc{Can}, we collect \textsc{Can} sentences among domains including talks, comments and dialogues from scratch.\footnote{\url{https://www.wikipedia.org}, \url{https://www.cyberctm.com}, \url{http://discuss.hk} and \url{https://lihkg.com}.}  
In order to maintain the consistency of training sets, \textsc{Man} corpora are also derived from same domains as \textsc{Can} from ChineseNlpCorpus and Large Scale Chinese Corpus for NLP.\footnote{\url{https://github.com/brightmart/nlp_chinese_corpus} and \url{https://github.com/SophonPlus/ChineseNlpCorpus}.}

\begin{table}[t]
\centering
\begin{tabular}{c| c c c c}
\hline
Dialect & \# Sents & Vocab size & Unique \\
\hline \hline
\textsc{Can} & 20M & 9,025 & 541 \\
\textsc{Man} & 20M & 8,856 & 372 \\
\hline
\end{tabular}
\caption{Statistics of two monolingual corpora after preprocessing. We conduct experiment at character-based level, and the joint vocabulary size is exactly 9,397.}
\label{Tab.CorporaStats}
\end{table}
\paragraph{Parallel Corpus}
We collect adequate parallel corpora for the development and evaluation of models. Parallel sentence pairs from dialogues are manually selected by native \textsc{Can} and \textsc{Man} speakers. Consequently, 1,227 and 1,085 sentence pairs are selected as development and test set, respectively.

\paragraph{Data Preprocessing \& Statistics}
As there is no well-performed \textsc{Can} segment toolkit, we conduct all the experiments at character level. 
In order to share the commonality of both languages and reduce the size of vocabularies, we convert all the texts into simplified Chinese.\footnote{We also attempt to transform all the texts into traditional characters. It does not work well since some simplified characters has multiple corresponding traditional characters and such kind of one-to-many mapping results in ambiguity and data sparsity.} 
For reasons of computational efficiency, we keep the sentences whose length lies between 4 and 32, and remove sentences composing characters with low frequencies. Finally, each of \textsc{Man} and \textsc{Can} monolingual training corpora consists of 20M sentences. The statistics of training set are concluded in Tab.~\ref{Tab.CorporaStats}.
As seen, \textsc{Can} has larger vocabulary size and more unique characters than \textsc{Man}. 
To identify the commonality and diversity of \textsc{Can} and \textsc{Man}, we compute the Spearman's rank correlation coefficient \cite{zhelezniak2019correlation} between two vocabulary rankings by their frequencies within each corpus. 
The coefficient score of two full vocabularies is $0.81$ ($p<0.001$), meaning that the overall relation is significantly strong. 
While the coefficient score of the 250 most frequent tokens is $0.26$ ($p<0.001$), indicating that the relation is significantly weak. 
These results cater to our hypothesis that dialects share considerate commonality with each other, but  possess diversity upon most frequent tokens.

\begin{table*}[t]
    \centering
    \begin{tabular}{l|c c c}
        \hline
        Model &  \textsc{Can}$\Rightarrow$\textsc{Man} & \textsc{Man}$\Rightarrow$\textsc{Can} & \# Params (M) \\ 
        \hline
        \hline
        \multicolumn{4}{c}{\textit{Baseline}} \\
        \hline
        Character-level Rule-based Transition & 42.18 & 42.27  & - \\
        Unsupervised Style Transfer \cite{hu2017toward} & 41.97 & 42.03 & 14.40 \\
        Unsupervised PB-SMT \cite{lample2018phrase} & 42.12  & 42.20  & - \\
        Unsupervised NMT \cite{lample2018phrase} & 42.90   & 42.39   & 39.08 \\
        \hline
        \hline
        \multicolumn{4}{c}{\textit{Ours}} \\
        \hline
        Layer Coordination & 48.45  & 43.11  & 39.08 \\
        Pivot-Private Embedding & 52.74  & 46.69   & 36.65 \\
        Pivot-Private Embedding + Layer Coordination & \textbf{54.95}  & \textbf{47.45}   & 36.65 \\
        \hline
        
    \end{tabular}
    
    \caption{Experimental results on unsupervised dialect neural machine translation. \# Params (M): number of parameters in million. We can see that layer coordination provides improvement over baseline on both directions, and pivot-private embedding improves the result further by almost 10 BLEU scores on \textsc{Can}$\Rightarrow$\textsc{Man}. Combining both layer coordination and pivot-private embedding gives the best result, exceeding 12 and 5 BLEU scores than baseline NMT system on two directions, respectively. }
    \label{Tab.AllResults}
\end{table*}

\section{Experiments}
\label{sec.Experiments}

\subsection{Experimental Setting}
\label{subsec.ModelTraining}

We use \textsc{Transformer}~\cite{vaswani2017attention} as our model architecture, and follow the base model setting for our model dimensionalities.
We refer to the parameter setting of ~\citeauthor{lample2018phrase}~\shortcite{lample2018phrase}, and implement our approach on top of their source code.\footnote{https://github.com/facebookresearch/UnsupervisedMT}
We use BLEU score as the evaluation metric. 
The training of each model was early-stopped to maximize BLEU score on the development set. 

All the embeddings are pretrained using fasttext \cite{bojanowski2017enriching},\footnote{https://github.com/facebookresearch/fastText} and pivot embeddings are derived from concatenated training corpora. 
In the procedure of training, $\lambda_{div}$ is set to 1.0, while $\lambda_{com}$ is linearly decayed from 1.0 at the beginning to 0.0 at the step being 200k.

\paragraph{Baseline}
We compare our model with four systems:
\begin{itemize}
    \item We collect simple \textsc{Can}-\textsc{Man} conversion rules and  regard character-level transition as one of our baseline systems.
    \item Our model is built upon unsupervised NMT methods, we choose one of the most widely used architecture \cite{lample2018phrase} as our baseline system.
    \item Moreover, unsupervised phrase-based statistical MT \cite{lample2018phrase} has comparable performance to its NMT counterpart. Therefore, we also take unsupervised PB-SMT model into account.
    \item For reference, we also examine whether a style transfer system \cite{hu2017toward} can handle dialect translation task.   
\end{itemize}

\subsection{Overall Performances}
\label{sec.Results}
Tab.~\ref{Tab.AllResults} lists the experimental results. As seen,
character-level rule-based translation system performs comparably with conventional unsupervised NMT system. 
This is in accord with~\citeauthor{lample2018phrase}~\shortcite{lample2018phrase} that training process of unsupervised NMT is vulnerable, because no aligned information between languages can be afforded to model training. Relatively, character transition rules offer adequate aligned references to conduct the fairish results. Besides, the unsupervised PB-SMT model performs slightly worse than NMT system, a possible reason is that the model is hard to extract a well-performed phrase table from colloquial data~\cite{laurens97lexicalist}. We also evaluate a style transfer system~\cite{hu2017toward}. The model underperforms unsupervised NMT baseline, indicating that, to some extent, style transfer is not adequate for dialect translation. 

As to our proposed methods, layer coordination improves the performance by more than 5 BLEU scores at \textsc{Can}$\Rightarrow$\textsc{Man} direction, proving that sharing coordinate information at the same semantic level among dialects is effective. Besides, using pivot-private embedding further gives a higher increase of nearly 10 BLEU scores as well as reducing the model size, verifying that jointly modeling commonality and diversity of both dialects is both effective and efficient. Furthermore, combining both of above can give us more than 12 BLEU scores improvement than baseline NMT system, revealing that both pivot-private embedding and layer coordination are complementary to each other. As to the \textsc{Man}$\Rightarrow$\textsc{Can} direction, we can also observe improvements of our proposed methods. Translating \textsc{Man} to \textsc{Can} is more difficult since it contains more one-to-many character-level transition cases than its reversed direction. Despite this, our best approach still gains 5 BLEU scores improvement than baseline systems on \textsc{Man}$\Rightarrow$\textsc{Can} translation, revealing the universal effectiveness of our proposed method.
\begin{table}[t]
\centering
\begin{tabular}{l| c c}
\hline
Model & \textsc{Can}$\Rightarrow$\textsc{Man}  & \textsc{Man}$\Rightarrow$\textsc{Can} \\
    \hline
    \hline
    Baseline & 1.80 $\pm$ 0.44 & 2.57 $\pm$ 0.50 \\
    Our Model & ~~~2.50 $\pm$ 0.87 $\uparrow$ & ~~~3.16 $\pm$ 0.61 $\uparrow$ \\
    \hline
\end{tabular}
\caption{Human assessment on our experimental results. $\uparrow$: improvement is strongly significant ($p<0.01$). }
\label{Tab.human}
\end{table}

\paragraph{Human Assessment}
Since BLEU metric may be insufficient to reflect the quality of oral sentences, we randomly extract 50 \textsc{Can} $\Rightarrow$ \textsc{Man} and 50 \textsc{Man} $\Rightarrow$ \textsc{Can} examples from test set for human evaluation, respectively. Each example contains source sentence, translated sentences from Unsupervised NMT model (``baseline'') and our proposed model. Each native speaker is asked to present a score ranging from 1 to 4 to determine the translation quality of each translated result within each example. Each of the reported result is the average score assessed by 10 native speakers. As seen in Tab.~\ref{Tab.human}, results prove that proposed method significantly outperforms baseline NMT system ($p<0.01$) in both \textsc{Can}$\Rightarrow$\textsc{Man} and \textsc{Man}$\Rightarrow$\textsc{Can} directions.

\subsection{Effectiveness of Pivot-Private Embedding}
\label{subsec.AnalysisPivot}

\begin{figure}[t!]
    \hspace{-5pt}
    \centering
    \includegraphics[keepaspectratio,height=0.55\columnwidth]{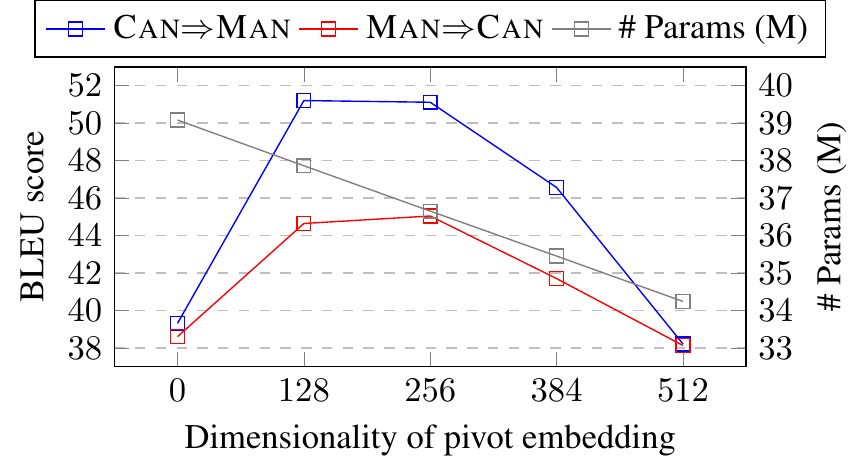}
    \caption{Model performances with various pivot embedding dimensionalities upon dev set. \# Params (M): number of parameters in million. We can observe that applying adequate dimensionality to pivot embedding is effective, rather than non-sharing any dimension among two dialects (dimensionality is 0) or sharing all dimensions (dimensionality is 512). }
    \label{Fig.PivotEmbedding}
\end{figure}

To investigate the effectiveness of pivot-private embedding, we also conduct further research on the dimensionality of pivot embedding. As seen in Fig.~\ref{Fig.PivotEmbedding}, adequately sharing part of word embedding among dialects can greatly improve the effect, while using two independent sets of embedding for dialects, or sharing all dimensions of embedding leads to poor results. This indicates the importance of balancing the commonality and diversity for dialect translation. Moreover, the more the dimensionalities assigned to pivot embedding, the fewer the parameters required by models. We argue that using pivot-private embedding is not only an efficient way to augment the ability of dialect translation system to model diversity, but also offer an alternative way to relieve the effect of over-parameterization. 

Comparing to the model with the dimensionality being 128, the model with 256 pivot embedding dimensions yields comparable results on the two translation directions, while assigns fewer parameters. Consequently, we apply 256 as our default setting for pivot embedding dimensionality.  

\subsection{Effectiveness of Layer Coordination}
\label{subsec.AnalysisCoCAN}
\begin{figure}[t!]
    \hspace{-5pt}
    \centering
    \includegraphics[keepaspectratio,height=0.55\columnwidth]{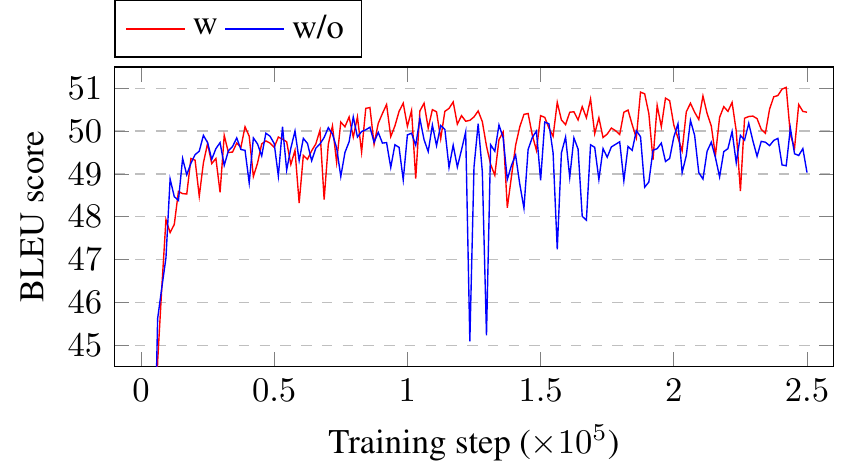}
    \caption{Learning curves of models upon dev set. Model with layer coordination (w) reaches its convergence at around step 240k, while model without (w/o) at around step 200k. As seen in this figure, applying layer coordination improves the performance of dialect translation model, as well as significantly stabilizes the training process.}
    \label{Fig.Convergence}
\end{figure}

\begin{figure*}[t!]
    \centering
    \hspace{-10pt}
    \begin{subfigure}[c]{\columnwidth}
    {
        \centering\includegraphics[keepaspectratio,width=0.85\columnwidth]{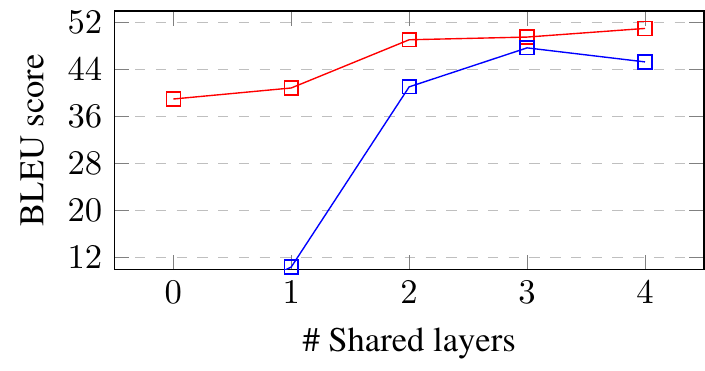}
        \caption{\textsc{Man}$\Rightarrow$\textsc{Can}}
    }
    \end{subfigure}
    \quad
    \hspace{15pt}
    \begin{subfigure}[c]{\columnwidth}
    {
        \centering\includegraphics[keepaspectratio,width=0.85\columnwidth]{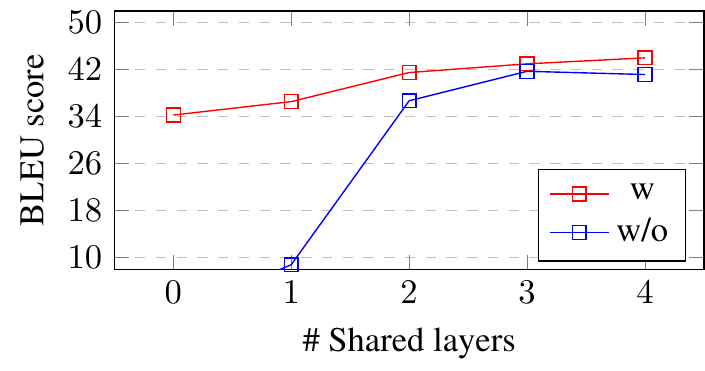}
        \caption{\textsc{Can}$\Rightarrow$\textsc{Man}}
    }
    \end{subfigure}

    \caption{Experiments on number of shared encoder/decoder layers upon dev set. Here w and w/o denotes with and without layer coordination, respectively. From both figures, we can see that even without any shared layer, model with layer coordination can also be trainable rather than without.  Models without layer coordination gain significant improvement upon sharing adequate layers for two dialects, while the performances decrease if all layers are shared. As to proposed layer coordination, the more layers shared for two dialects, the higher performance models can possess.}
    \label{Fig.Coor}
\end{figure*}

Layer coordination intuitively interacts features from all dialects, helping the model to capture the commonality of linguistic features at coordinate semantic level \cite{Peters:2018:NAACL}.  
\citeauthor{he2018layer}~\shortcite{he2018layer} reveal that layer coordination offers more aligned features at the same level, from lexical, through syntactic, to semantic. In this section, we investigate how layer coordination effects on translation quality.

\paragraph{Stability Analysis}
We first visualize the convergence of models with and without layer coordination. From Fig.~\ref{Fig.Convergence} we can observe that the model with layer coordination gains a steady training process, whereas training process of model without layer coordination is fragile, especially drop nearly 5 BLEU scores upon dev set at the middle term. 
We attribute this to the fact that layer coordination provides coordinate semantic information~\cite{he2018layer}, which is beneficial to our dialect translation task with respect to commonality modeling. Since the two dialects share similar features, each decoder layer can leverage more fine-grained  information from source side at the same semantic level, instead of only exploiting top-level representations.

\paragraph{Parameter Sharing}
For further investigation, we also conduct analyses on the effect of shared layers.
As visualized in Fig.~\ref{Fig.Coor}, baseline system performs worse when the number of shared layer is less than 1, and models with 3 layers shared performs better.  This is consistent with findings in \citeauthor{lample2018phrase}~\shortcite{lample2018phrase} who suggest to share higher 3 layers in encoder and lower 3 ones in decoder. 
Considering the proposed model, sharing more layers for \textsc{Can} and \textsc{Man} translation on both directions is profitable, and model with all layers shared gives the best performance on both directions. This demonstrates that \textsc{Can} and \textsc{Man} have more similar characteristics in numerous aspects of linguistics than distant languages \cite{artetxe2018unsupervised,lample2018unsupervised}, and layer coordination also contributes to the balance of commonality and diversity modeling upon dialect translation task.

\section{Related Work}
\label{sec.RelatedWork}

In this section, we will give an account of related research.
\paragraph{Dialect Translation} To the best of our knowledge, related studies on dialect translation have been carried upon a lot of languages. For example, in Arabic \cite{baniata2018neural} and Indian \cite{chakraborty2018bengali}, applying syllable symbols is effective for sharing information across languages. Compared to these tasks, our work mainly focus on handling problems in \textsc{Can} and \textsc{Man} translation task.  \textsc{Can} and \textsc{Man} have little syllable information in common, as even the same character can  be widely divergent in aspect of pronunciation~\cite{lee1998cancorp,wong2018register}. To push the difference further, a set of \textsc{Can} characters is quite rarely to be seen in \textsc{Man}, because \textsc{Can} is a dialect that without formal regulation of written characters~\cite{lee1998cancorp}. Moreover, younger \textsc{Can} speakers more likely refer to use phonetic labels (e.g. ``d'' responses to ``di'') or homophonetic character symbols instead of ground truth, which raises intractable issues when building the translation model. 

\paragraph{Unsupervised Learning}
Our work refers to quantitative researches on unsupervised machine translation \cite{lample2018unsupervised,artetxe2018unsupervised,lample2018phrase}, which compose a well-designed training schedule for unsupervised translation tasks. 
The difference between our research and theirs mainly lies in the similarity of involved languages, where dialects in our research are far similar with each other than those in unsupervised NMT tasks. 

Moreover, our research is closely related to studies on style transfer~\cite{hu2017toward,prabhumoye2018style}. There are two main differences between our task and style transfer. Firstly, the source and target sides in style transfer task belong to the same language, where the difference mainly contributed by style, e.g. sentiment~\cite{hu2017toward}, while dialect translation has to identically guarantee the semantics between two sides. Secondly, there are more commonalities between source and target in style transfer than that in dialect translation. The former focus on the transition of different styles, the two sides can sometimes be distinguished by only a few words. Nevertheless, dialects have wide discrepancies which vary from vocabulary and word frequency to syntactic structure.

Methodologically, compare to the mentioned studies, we motivated by similarity and difference between dialects and propose pivot-private embedding and layer coordination to jointly balance {\em commonality} and {\em diversity}.

\section{Conclusions and Future Work}
\label{sec.Conclusion}

In this study, we investigate the feasibility of building a dialect machine translation system.
Due to the lack of parallel training corpus, we approach the problem with unsupervised learning. Considering the characteristics in dialect translation, we further improve our translation model by contributing pivot-private embedding and layer coordination, thus enriching the mutual linguistic information sharing across dialects (\textsc{Can}-\textsc{Man}). Our experimental results confirm that our improvements are universally-effectiveness and complementary to each other. 
Our contributions are mainly in:
\begin{itemize}
    \item We propose dialect translation task, and conduct massive examples of monolingual sentences with respect to dialects of spoken \textsc{Man} and \textsc{Can};
    \item We apply an unsupervised learning algorithm to accomplish \textsc{Can}-\textsc{Man} dialect translation task. We leverage {\em commonality} and {\em diversity} modeling to strengthen the translation functionality among dialects, including pivot-private embedding and layer coordination;
    \item Our approach outperforms conventional  unsupervised NMT system over 12 BLEU scores, achieving a considerable performance and a new benchmark for the proposed \textsc{Can}-\textsc{Man} translation task. 
\end{itemize}

In the future, it is interesting to validate our principles, i.e. commonality and diversity modeling, into other tasks, such as conventional machine translation and style transfer. 
Another promising direction is to incorporate linguistic knowledge into unsupervised learning procedure, e.g. phrasal pattern~\cite{xu19leveraging}, word order information~\cite{yang19assessing} and syntactic structure~\cite{yang2019improving}. 

\section{Acknowledgements}
This work was supported in part by the National Natural Science Foundation of China (Grant No. 61672555), the Joint Project of the Science and Technology Development Fund, Macau SAR and National Natural Science Foundation of China (Grant No. 045/2017/AFJ), the Science and Technology Development Fund, Macau SAR (Grant No. 0101/2019/A2), and the Multi-year Research Grant from the University of Macau (Grant No. MYRG2017-00087-FST). We thank all the reviewers for their insightful comments.

\bibliographystyle{aaai}
\bibliography{all}

\end{document}